\documentclass[journal]{IEEEtran}
\DeclareUnicodeCharacter{2113}{l}
\usepackage{amsmath, amssymb}
\usepackage{url}
\usepackage[utf8]{inputenc}
\usepackage[T1]{fontenc}
\usepackage{textcomp}
\usepackage{tabularx}
\usepackage{floatrow}
\usepackage{graphicx}
\usepackage{comment}
\usepackage[skip=2pt,font=small]{caption}
\usepackage{color,soul}
\usepackage{subcaption}
\usepackage{float}
\usepackage{multirow}
\usepackage{multicol}

\definecolor{Gray}{gray}{0.85}
\definecolor{LightCyan}{rgb}{0.88,1,1}
\makeatletter

\begin{document}

\title{Deep Representation for Connected Health: Semi-supervised Learning for Analysing the Risk of Urinary Tract Infections in People with Dementia}

\author{Honglin Li,
        Magdalena Anita Kolanko,
        Shirin Enshaeifar,
        Severin Skillman,
        Andreas Markides,
        Mark Kenny,
        Eyal Soreq,
        Samaneh Kouchaki,
        Kirsten Jensen,
        Loren Cameron,
        Michael Crone,
        Paul Freemont,
        Helen Rostill,
        David J. Sharp,
        Ramin Nilforooshan,
        Payam Barnaghi
\thanks{H. Li, M. N. Kolanko, S. Skillman, E. Soreq, D. J. Sharp and P. Barnaghi are with Department of Brain Sciences, Imperial College London, W12 0NN, United Kingdom.}
\thanks{S. Enshaeifar, A. Markides and S. Kouchaki are with Centre for Vision, Speech and Signal Processing, University of Surrey, Guildford, GU2 7XH, United Kingdom.}%
\thanks{K. Jensen, L. Cameron, M. Crone and P. Freemont are with Section of Structural and Synthetic Biology, Department of Infectious Disease, Imperial College London, W2 1NY, United Kingdom.}
\thanks{R. Nilforooshan, M. Kenny and H. Rostill are with Surrey and Borders NHS Foundation Trust, Leatherhead, KT22 7AD, United Kingdom.}
\thanks{All authors are also with the Care Research and Technology Centre, The UK Dementia Research Institute (UK DRI).}
}

\maketitle
\IEEEpeerreviewmaketitle 
 \begingroup\renewcommand\thefootnote{\textsection}
\footnotetext{Corresponding author: p.barnaghi@imperial.ac.uk}
\endgroup

\begin{abstract}
Machine learning techniques combined with in-home monitoring technologies provide a unique opportunity to automate diagnosis and early detection of adverse health conditions in long-term conditions such as dementia. However, accessing sufficient labelled training samples and integrating high-quality, routinely collected data from heterogeneous in-home monitoring technologies are main obstacles hindered utilising these technologies in real-world medicine. This work presents a semi-supervised model that can continuously learn from routinely collected in-home observation and measurement data. We show how our model can process highly imbalanced and dynamic data to make robust predictions in analysing the risk of Urinary Tract Infections (UTIs) in dementia. UTIs are common in older adults and constitute one of the main causes of avoidable hospital admissions in people with dementia (PwD). Health-related conditions, such as UTI, have a lower prevalence in individuals, which classifies them as sporadic cases (i.e. rare or scattered, yet important events). This limits the access to sufficient training data, without which the supervised learning models risk becoming overfitted or biased. We introduce a probabilistic semi-supervised learning framework to address these issues. The proposed method produces a risk analysis score for UTIs using routinely collected data by in-home sensing technologies. The solution is designed utilising a dataset collected in a remote healthcare monitoring study of people affected by dementia with 110 participants with a mean age of 83 and a standard deviation of 6 (57 males and 53 females). The data was acquired from a set of network-connected sensory devices deployed in patients' homes. The design and validations of the model were conducted on a dataset containing 3,864 days of patient data. To evaluate the generalisation and scalability of the model, we performed further evaluations with an additional dataset from our study that includes 13,090 days of data from 68 new patients (32 males and 36 females) with a mean age of 81 and a standard deviation 13.7. The proposed machine learning model can detect the risk of UTI with a precision of 86\%, recall of 85\% using in-home monitoring data. The results validated in a clinical study by a monitoring team who contact the patient or carer when a UTI alert is generated to evaluate the symptoms and then follow up the GP visit and urine test results. 
\end{abstract}

\section{Introduction}
Urinary Tract Infections (UTIs) are the most common bacterial infection in elderly patients \cite{linhares2013frequency}, and one of the top causes of hospital admission in people with dementia (PwD) \cite{rao2016outcomes,masajtis2017new,sampson2009dementia, burnham2018urinary}. In the UK, UTIs contribute to over 8\% of the hospital admissions in PwD while at any given time, 1 in 4 of the hospital beds are occupied by someone with dementia. Nevertheless, diagnosis of UTI in elderly patients remains difficult due to the high prevalence of asymptomatic bacteriuria (AB)~\cite{gavazzi2013diagnostic,nicolle2003asymptomatic} and presence of a range of nonspecific (or atypical) symptoms~\cite{marques2012epidemiological}. These factors delay diagnosis of UTI, and the condition often remains undiagnosed until it progresses to severe symptoms or complications requiring hospital admission~\cite{gavazzi2013diagnostic,chu2018diagnosis}. UTI is also the most common cause of sepsis in older adults \cite{peach2016risk} with an associated in-hospital mortality of 33\% in this age group \cite{tal2005profile}. It has been demonstrated that a delay in UTI treatment is associated with an increased risk of bloodstream infection and all-cause mortality compared with immediate treatment \cite{gharbi2019antibiotic}. Currently, urine and blood tests are standard UTI diagnostic methods. However, these methods are time consuming, and more importantly, they are only utilised once there is clinical suspicion of a UTI. Therefore, highlighting the ``possibility'' of UTI based on early symptoms is very important for older people as it allows timely UTI detection and intervention, ultimately preventing unnecessary hospital admissions and UTI complications \cite{chu2018diagnosis, marques2012epidemiological}. 

Early detection of UTI is particularly important for PwD who might have reduced or delayed help-seeking behaviour, either through reduced recognition of symptoms or impaired communication skills. Consequently, an acute infection might not be diagnosed in PwD until physical symptoms become severe or secondary behavioural/psychological symptoms of delirium develop \cite{toot2013causes}. 
Recent advances in machine learning show promise for development of more sophisticated clinical decision support tools with predictive models that incorporate dynamic healthcare data to improve diagnostic accuracy. While such predictive models have been proposed for UTI risk prediction, they are limited by small data sets, poor generalisability to an elderly population, and insufficient diagnostic performance \cite{little2006developing, mcisaac2007validation, heckerling2007predictors, papageorgiou2012fuzzy}. Furthermore, they frequently rely on the presence of typical symptoms, as well as urinalysis and laboratory findings as predictor variables, precluding their wider use in community-dwelling patients with dementia, who often present atypically and may have delayed access to blood or urine testing.  

‘Smart-home’ technologies offer an alternative approach to remote healthcare monitoring and provide novel opportunities for detecting clinically significant events through the use of wearable technology, physiological monitoring devices or environmental sensors \cite{majumder2017smart, turjamaa2019smart, peetoom2015literature}. Computational models have been developed to analyse sensor data and identify changes in activity patterns over time to predict deterioration in health status. Much of the previous work focused on detection and prediction of falls \cite{schwickert2013farseeing}, as well as tracking behavioural symptoms such as sleep disturbances \cite{lazarou2016novel}, agitation \cite{bankole2012validation}, and wandering \cite{fleiner2016sensor} in elderly patients. However, there is limited research on the use of machine learning models for detection of infection in the dementia population in the context of ‘smart-home’ technologies. Rantz et al. \cite{rantz2011using}, have previously developed a solution for detecting UTI symptoms using in-home passive infra-red (PIR) sensory data. However, their method uses a supervised learning approach which depends on the activity labels and annotations in the training dataset. The activity labelling in their work is performed based on a manual annotation procedure, which is an extremely time-consuming task and difficult to scale-up, limiting the generalisability of their algorithm to real-world scenarios. In an early work, we used an supervised learning model to cluster patient data to UTI/Non-UTI days \cite{enshaeifar2019machine} (i.e. similar to zero-shot learning models). However, this method had a very low precision and did not generalise when the data changed over time and also when the trained model was tried with a different population. This led to our new research to develop solution than can utilise both unlabelled and labelled data in a semi-supervised model and can generalise when the population or data distribution changes.    

\begin{figure*}[t]
		\centering
		\includegraphics[width=0.80\textwidth]{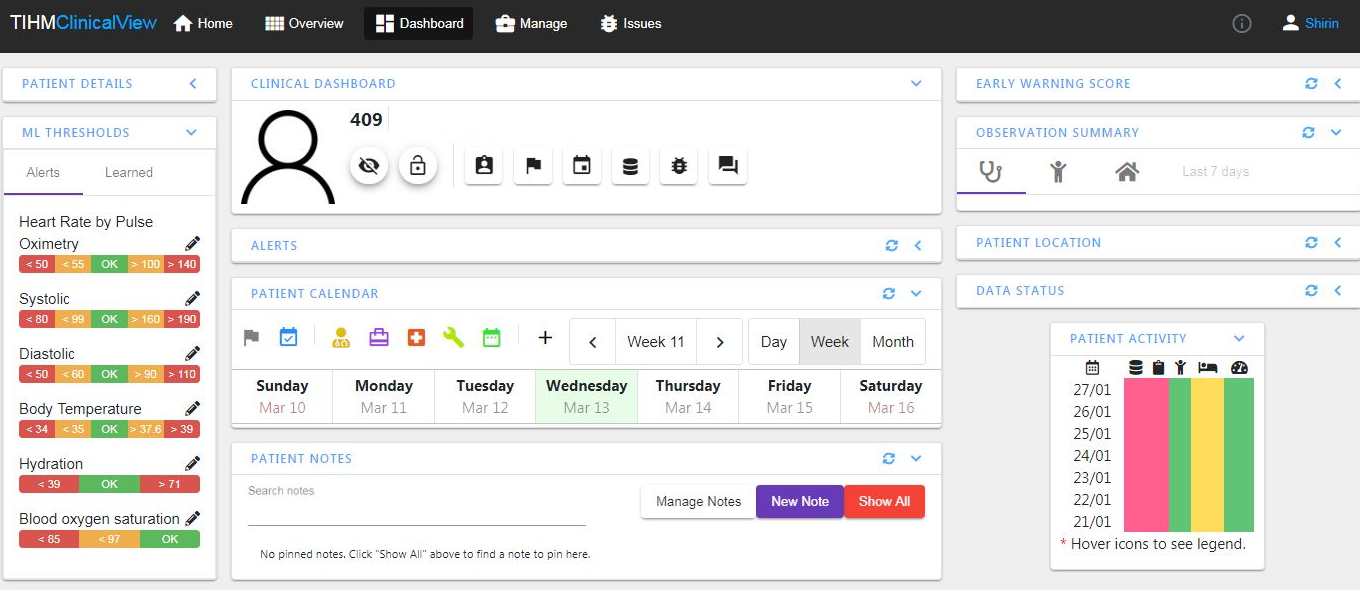}
		\caption{A screenshot of the dashboard of the developed digital platform, used  by the study's clinical monitoring team for responding to machine generated alerts and to provide continuous health monitoring.
} 
		\label{fig:iview}
\end{figure*}

Our new approach for analysing the risk of UTIs in PwD relies on a unique technology-assisted monitoring system, which uses off-the-shelf and low-cost in-home sensory devices to continuously monitor environmental and physiological data of PwD within their own homes~\cite{enshaeifar2018health,enshaeifar2018UTI}. By fusing novel machine learning and analytic approaches, we analyse the combination of environmental and physiological data to extract and identify early symptoms and the risk of UTI. A digital platform we designed for data collection and for the integration of machine learning algorithms has been licensed as a Class-I medical device (see Fig. \ref{fig:iview}). We have also developed clinical pathways to respond to the risk alerts generated by our analytical algorithms. 

The network-connected sensory devices have played an important role in the development of remote monitoring systems for healthcare applications~\cite{doukas2012bringing,yang2014health,catarinucci2015iot,amendola2014rfid}; however, training machine learning models for the sensory data still poses significant challenges: (i) the sensory data collected from real-world environments are noisy, incomplete (containing missing values) and unreliable due to uncontrolled recording conditions. Effective pre-processing techniques are required to enhance data quality where needed (ii) streaming sensory data from real-world environments generates large volumes of continuous data. This requires scalable and adaptive algorithms for analysing time-series data while identifying and analysing sporadic events such as UTIs (i.e. rare or scattered, yet important events) (iii) real-world data is dynamic and often change over time due to environmental, activity or health variations. Hence, new solutions are required to create adaptive algorithms for analysing the continuous data.  

To address the challenges mentioned above, we propose a semi-supervised approach which leverages the advantages of adaptive Deep Neural Network (NN) algorithms and supervised probabilistic models. A variety of semi-supervised learning methods have been previously developed~\cite{coates2011analysis,sheikhpour2017survey}. For example, generative models utilise unlabelled data to identify the mixture distribution and the labelled data to determine the class~\cite{zhu2006semi}; self-training methodologies classify the unlabelled data and re-train the data repeatedly~\cite{prakash2014survey}; graph-based models define the relationship between labelled and unlabelled data by a graph~\cite{zhu2009introduction}. In general, semi-supervised learning approaches have been widely utilised for datasets with limited labelled data~\cite{zhu2003semi}. 

We propose a semi-supervised approach which first trains a model on the unlabelled data using unsupervised techniques and then applies the trained model on the smaller set of labelled data to develop a supervised classifier with high reliability. We use probabilistic and deep learning models to analyse the risk of UTIs using environmental and physiological observations. In our previous work~\cite{enshaeifar2018UTI}, we developed an unsupervised algorithm for detecting the risk of UTIs from unlabelled data, and manually validated the identified UTI events. Due to the lower prevalence of UTIs in individuals, our previous algorithm  only had an accuracy of 14\%~\cite{enshaeifar2018UTI}. In this study, we propose a semi-supervised model which significantly improves the validation rate to 85\%.

The clinical design and workflow proposed by this work offer new pathways for remote monitoring and care technologies to support PwD, which is generalisable to other long-term conditions. The proposed machine learning model is currently deployed in our digital platform \cite{TIHM2020}, which is a CE marked Class-I medical device. In terms of the technical contributions, this work proposes a continual learning method that shows strong performance in handling partially labelled and imbalanced data.

The model is generalisable to new training samples and can continuously learn by observing new labelled samples. In terms of the application contribution, this work demonstrates how low-cost in-home sensing technologies can be utilised to effectively monitor chronic health conditions. The digital platform presented in this paper has been tested in an observational clinical study and deployed in 143 homes of people living with dementia.

\section{Material and Methods}
This section describes the proposed semi-supervised model and the data collection procedure used in the study \footnote{The study has received ethic approval from South East Coast Surrey NHS Research Ethics Committee (i.e. SURREY, NRES Committee SE Coast (HEALTH RESEARCH AUTHORITY); TIHM REC: 16/LO/1802; IRAS: 211318).}. 

\subsection{The Semi-supervised Model} 
The continuous data collected from the real-world is often dynamic, and changes due to environmental factors or patient health variations. To develop a continually learning model, we use supervised neural network which fit a model without assuming a prior distribution. However, these algorithms require a large amount of labelled data to be trained with all the variations or expected tasks~\cite{schmidhuber2015deep}. Labelling the data (especially those collected from uncontrolled environments or those including sporadic cases) is an expensive, time consuming and challenging process. Therefore, the supervised NN algorithms are not optimal solutions for dealing with datasets with insufficient labelled examples. On the other hand, some of the conventional classifiers, such as Na\"ive Bayes and Logistic Regression algorithms, can converge with a small amount of labelled data; however, these algorithms require prior knowledge and regularisation that will affect their performances in changing environments.

\begin{figure}[t!]
\centering
\includegraphics[width=\linewidth, height=0.4\linewidth]{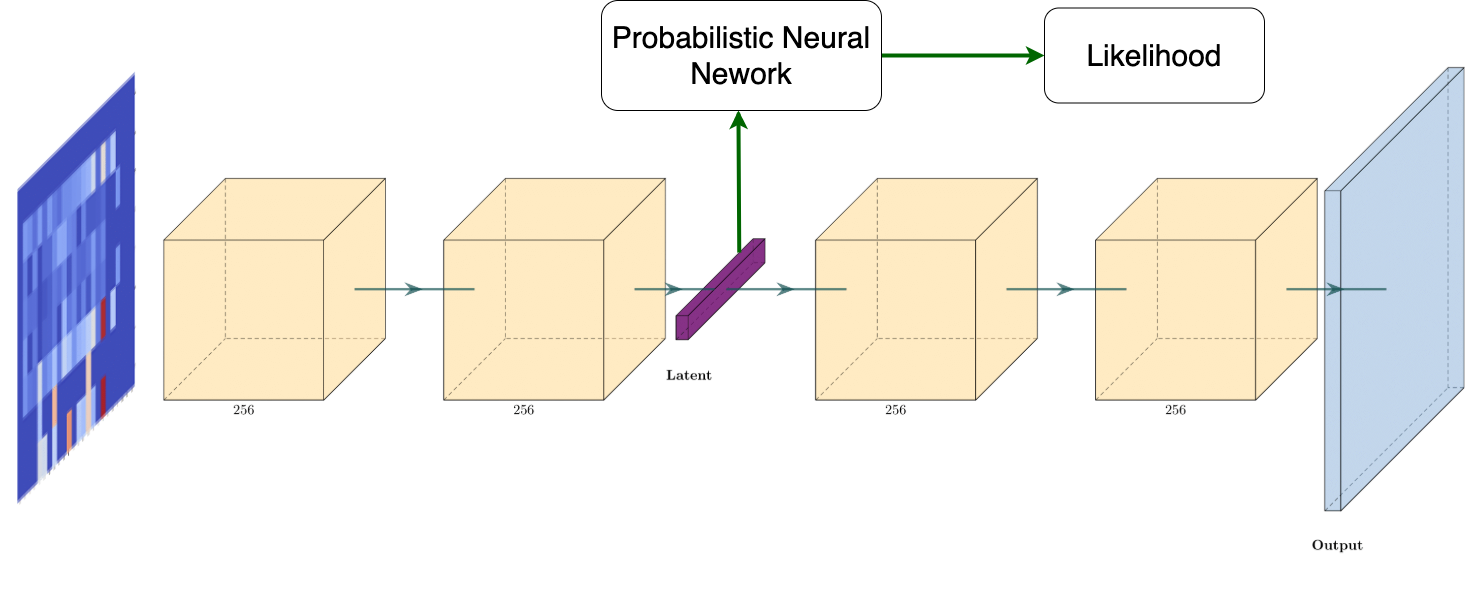}
\caption{An overview of the semi-supervised model, combining an auto-encoder with a deep probabilistic neural network. }
\label{fig:semi-sup}
\end{figure}

To leverage the advantages of both NN algorithms and probabilistic models and to include unlabelled data in the learning, we propose a semi-supervised learning framework~\cite{coates2011analysis} which combines unsupervised and supervised learning techniques. In this regard, a deep NN algorithm is used as an auto-encoder to learn a representation of the unlabelled data. In other words, the NN model performs as an unsupervised feature extractor which learns the intrinsic relations from the unlabelled data and creates a suitable model to represent the original data in a latent feature space. In the next step, the encoding model is applied to the smaller set of labelled data to extract the corresponding features, and they are then used to train a supervised classifier. An overview of the model is shown in Fig. \ref{fig:semi-sup}

By using an unsupervised feature extractor, the model is trained with a large volume of unlabelled data. This improves the performance of the supervised classification method in the second step. This is due to the fact that in training a model with smaller datasets, noise and bias can significantly affect the performance of the model; while this effect is attenuated by using a larger dataset in the initial feature extraction step. This semi-supervised model can be used to identify sporadic cases (such as UTIs or other healthcare conditions with low prevalence) even when there is a limited number of labelled examples within a large set of sensory data.

\begin{figure}[t!]
\centering
\includegraphics[width=\linewidth]{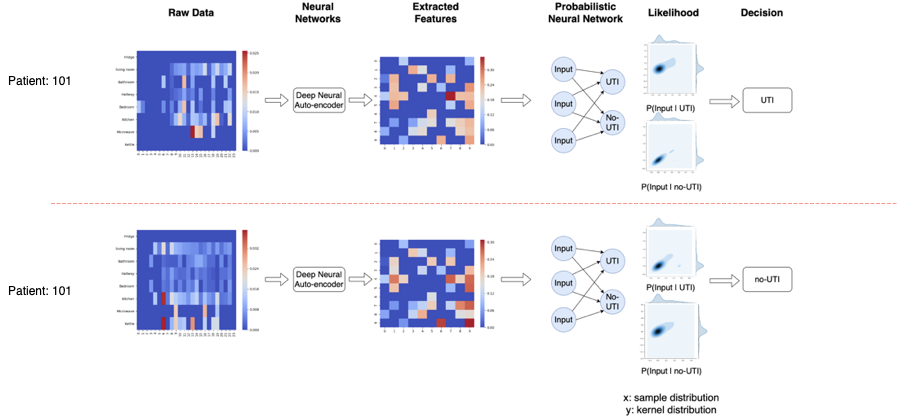}
\caption{An overview of the semi-supervised probabilistic neural network. The network can estimate the probabilistic density of the extracted features. If the distribution resides in the middle point of the y axis, the likelihood is high.}
\label{fig:pnn_example}
\end{figure}

\subsubsection{Auto-encoders}
Auto-Encoders (AE) consist of two layers referring to as encoder $\mathcal{G}$ and decoder $\mathcal{D}$~\cite{le2015tutorial}. Suppose that there is a set of training samples $\mathbf{X}=\{\mathbf{x}_1,\mathbf{x}_2,\dots,\mathbf{x}_k\}$. By setting the output values to be same as inputs, an auto-encoder aims to learn the function $H_{w,b}(\mathbf{x}_i) \approx \mathbf{x}_{i}$ where $H_{w,b}(\mathbf{x}_i) = \mathcal{D}(\mathcal{G}(\mathbf{x}_i))$. By using this training process, we can use labelled data to extract the latent features. Hence, the features extracted from the data will be $\mathbf{C}=\text{f}(x_i)$ referring to the coding process. The coded data $\mathbf{C}$ will then be used as the input of selected classifiers. 

In Convolutional Auto-encoders (CAE) are similar to other AE, except they contain several filters which share parameters and utilise the spatial correlations. CAEs have been widely used as a feature extractor technique~\cite{holden2015learning} in image classification. Similarly, we use convolutional layers to extract the underlying features from a set of unlabelled data (non-image matrices). 

\subsubsection{Probabilistic Neural Network Classifier}
We combine a Probabilistic Neural Network (PNN) \cite{specht1990probabilistic,li2020continual} with a fully connected neural network to classify the data based on the extracted latent features from the auto-encoder. Without the back-propagation training procedure, the PNN can be trained quickly but suffers from the massive growth of the number of nodes and parameters. However, PNNs can estimate the probabilistic density of the input and continuously learn from data by adding new kernels without suffering from the forgetting problem \cite{mccloskey1989catastrophic,li2020continual}. The model estimates the confidence of the predictions and provides continual learning for the new samples after the initial training process. 

We train the PNN with back-propagation and update the output nodes as new samples are observed. We stack an encoder onto the PNN to extract the latent features. Fig.~\ref{fig:pnn_example} shows an example of the model with different stages of data processing. Different from existing semi-supervised learning approaches, the training procedure is similar to the adversarial training. We train an auto-encoder and a PNN simultaneously to increase the margin between positive and negative samples. Utilising the PNN, we can estimate the density of the samples and the encoder will have an agreement with mapping the labelled samples and unlabelled samples into similar distributions. In summary, the PNN takes the output of the encoder as an input and follows the following proposed training procedure:
At the first stage, the auto-encoder will be trained to reconstruct the input samples. At the second stage, the PNN acts as a density estimator to estimate the probabilistic density of the labelled samples. 

PNN can estimate the probability density of the samples, and output the likelihood of the sample against each class, see Eq.~(\ref{eq:density_estimation}). The $x$ is the training samples, $g(x)$, $K$ is the kernel in the PNN, $\sigma$ is the standard deviation in PNN. $K$ and $\sigma$ are trainable parameters. If there are multiple kernels in PNN, the final probability will be calculated by the second equation in Eq.~(\ref{eq:density_estimation}). 

\begin{equation}
\label{eq:density_estimation}
\begin{split}
    \begin{aligned}
    & \varphi(x,K) = \exp\{\frac{||x - K||^2}{2 * \sigma^2}\},\\
    \quad
    & P = \frac{ \sum_i^n \varphi(x,c_i) }{\sum_i^n \varphi(x,c_i) + n - max \{\varphi(x,c_j):j=1,\dots,n\} * n}
    \end{aligned}
\end{split}
\end{equation}

We evaluated our new method against other existing solutions. To compare the performance of CAE with other techniques, we used Auto-encoder (AE), Deep-encoder (DE), and Restricted Boltzmann Machine (RBM) to represent the unlabelled data. Furthermore, we used a set of common methods including Logistic Regression (LR), Support Vector Machine (SVM) (with RBF and Polynomial kernels), Decision Tree (DT) and K-nearest Neighbour (KNN) algorithms to classify the data and provided numerical assessments (see the result section). The proposed method is validated in a real-world healthcare application and an online dashboard to generate and validate alerts indicating the risk of UTIs for PwD. 

\subsection{Data Collection and Variables} \label{sec var}
\begin{figure}[t!]
\centering
\includegraphics[width=0.5\linewidth]{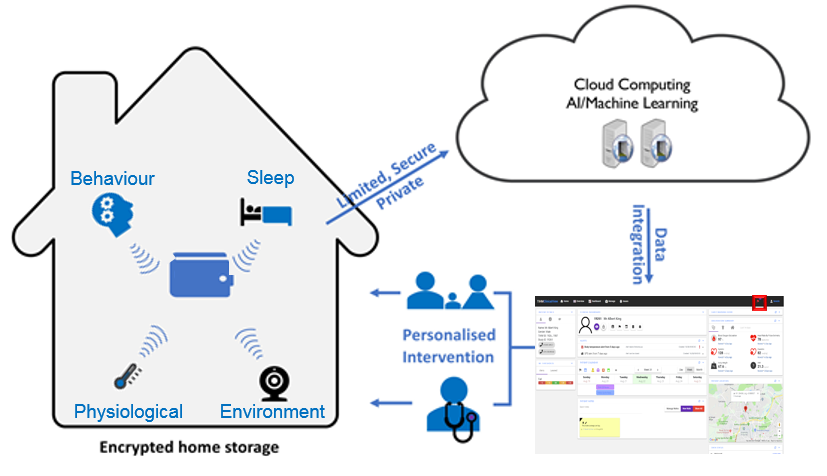}
\caption{High-level overview of the monitoring system installed in homes of people with dementia}
\label{fig:tihm}
\end{figure}

In our study, the environmental data were continuously recorded via sensors installed in participants' homes, see Fig.~\ref{fig:tihm}. The environmental sensors consist of passive infra-red (PIR) sensors (installed in the hallway and living room), movement sensors (one in the kitchen, and two on the bedroom and bathroom doors), pressure sensors (deployed on the bed and the chair), main entrance door sensor and smart plugs for appliance usage monitoring. The physiological data were recorded once a day by participants using Bluetooth-enabled medical devices. This includes blood pressure, heart rate, body temperature, weight and hydration readings, see Table \ref{tab:tab5}.

\begin{table*}[t]
    \centering
    \caption{Digital markers in in-home monitoring in dementia care }
    \small
    \begin{tabular}{*{4}{p{.23\linewidth}}}
        \textbf{Digital marker} & \textbf{Monitoring Device}   & \textbf{Frequency} &\textbf{Additional Information}
\\
\hline
\\
  Movement & Door, Motion and Tracking sensors  & Continuous (multiple doors, hallway, kitchen and rooms & Daily living activity pattern analysis and entropy measurement \cite{enshaeifar2018health}; Movement and gait speed \cite{van2006frail, castell2013frailty, grande2019measuring, shimada2018cognitive}; UTI and Agitation \cite{enshaeifar2019machine, enshaeifar2018health}.
 \\    
 \\
 Home Device Usage & Home appliance use, smart plugs & Continuous (kitchen appliances, TV and commonly used devices)  &  Daily living activity pattern analysis and entropy measurement \cite{enshaeifar2018health, daly2000predicting}; Measuring the correlation between Cognitive decline and home appliance usage \cite{hoang2016effect, fancourt2019television}.     
   
 \\   
 Body temperature & Smart Temporal Thermometer & Twice daily (morning, evening) or continuous using a wearable device & Infections, UTIs \cite{enshaeifar2019machine}.  
\\    
    
 Blood Pressure & BPM Connect or using a wearable device &  Twice a day (morning, evening) or frequent times daily & Risk analysis \cite{sibbett2017risk, gilsanz2017female}.    
 \\     
    
Walking (step count) & Step watch  & Continuous & Daily living activity pattern analysis and entropy measurement \cite{enshaeifar2018health, daly2000predicting}.    
\\      

Pulse & Pulse HR  & Continuous & Risk analysis \cite{sibbett2017risk, gilsanz2017female}.    
    
      
            
    \end{tabular}
    \label{tab:tab5}
  \end{table*}

\begin{figure}[t!]
\centering
\includegraphics[width=0.45\linewidth]{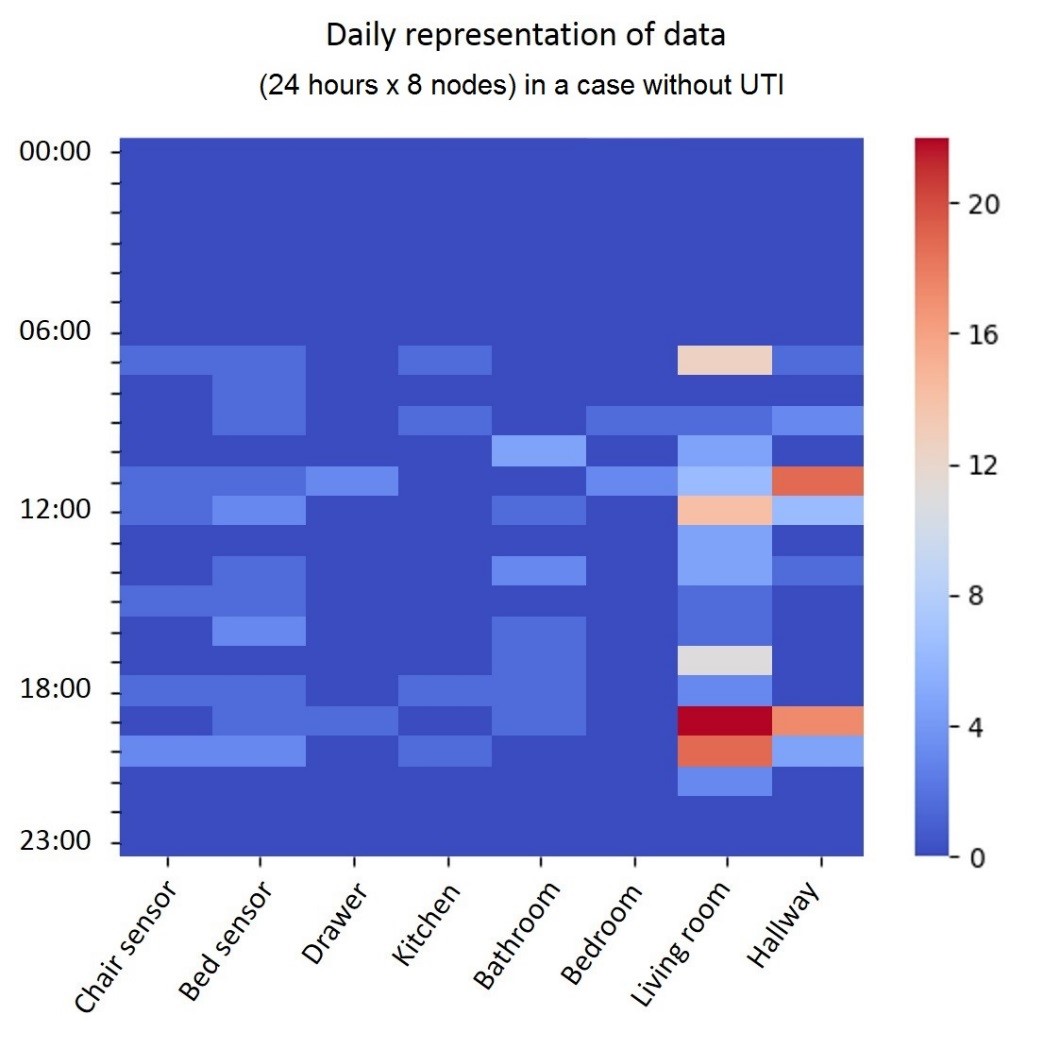}
\includegraphics[width=0.45\linewidth]{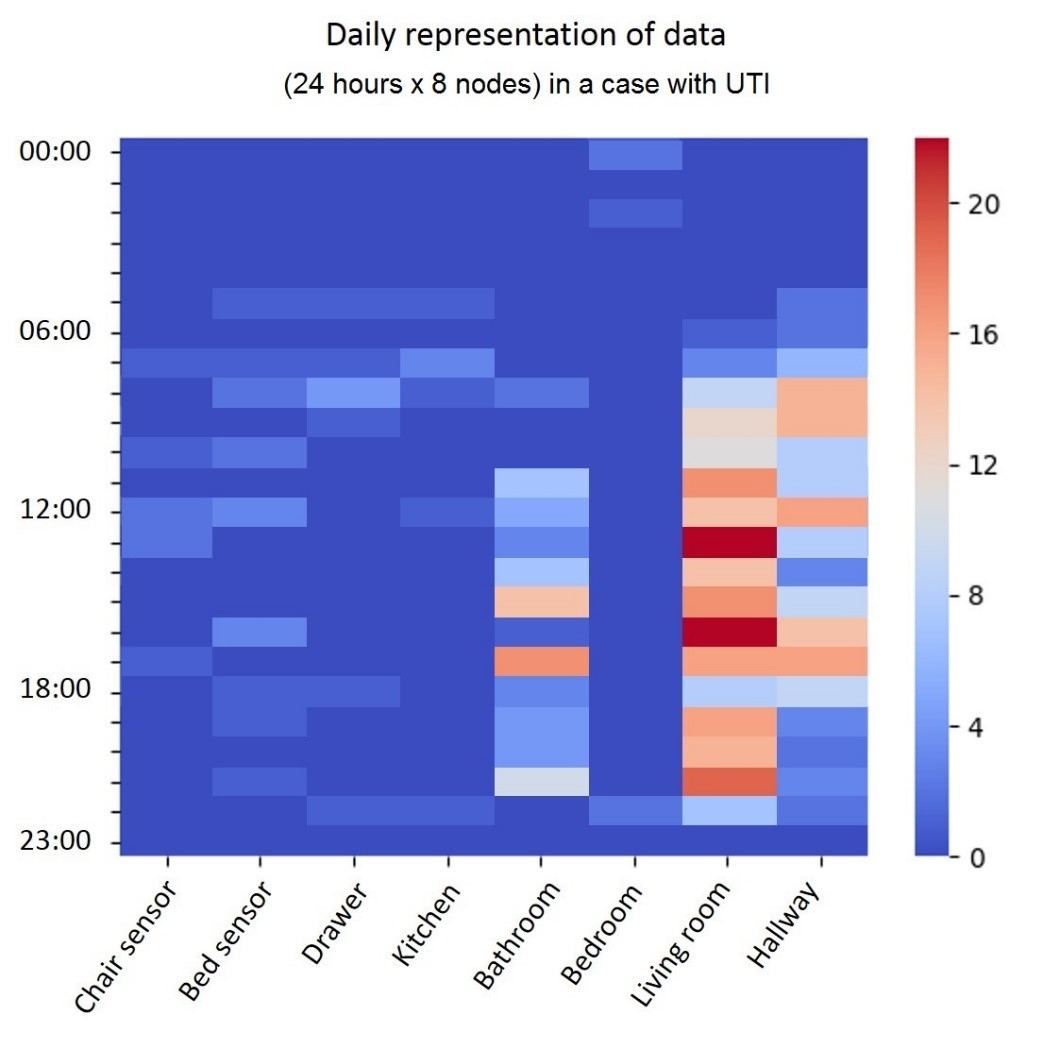}
\caption{Illustration of environmental data collected in two different days, one in a non-UTI day (left) vs. a UTI day (right). The environmental data is aggregated within hourly intervals for both days. The day with UTI (right) has an increased number of bathroom readings and it shows some movement activities throughout the night (00:00 to 06:00).}
\label{fig:daily}
\end{figure}

To analyse the collected data, we selected data for the environmental sensors. The reading are aggregated within hourly intervals to generate an accumulated Sensor Firing Pattern (SFP); i.e. each day is represented as a $24x8$ input matrix where $8$ was the number of nodes, see Fig.~\ref{fig:daily}. The hourly SFPs are then grouped to be used as the daily input data, and each day is labelled either as UTI or non-UTI.

The existing research shows that (i) UTIs increase frequency of bathroom usage during the day and night; (ii) more frequent bathroom visits at night could affect the quality of sleep, causing sleep disturbance and night-time wandering~\cite{agata2013challenges,juthani2007diagnostic}; (iii) UTIs in Pwd frequently lead to delirium, causing a significant and distressing change in their behaviour. These may also worsen pre-existing confusion and agitation in this population~\cite{agata2013challenges,juthani2007diagnostic,gavazzi2013diagnostic, ijaopo2017dementia}, making the patients deviate from their day-to-day activity patterns. Using environmental sensors such as movement sensors in bathrooms, pressure mats in bed, movement and door sensors around the house we remotely monitor bathroom usage, sleep routine and day-to-day activity patterns (including wandering at night-time). By continuously monitoring and analysing the environmental data, we can identify some of the symptoms mentioned above, which are associated with UTI, as shown in Fig.~\ref{fig:daily}.

To validate the proposed method, we compared the automatically generated labels against the manually annotated labels recorded by a clinical team during the study. In the study, we collect labelled data regarding the UTI events for our participants by taking urine samples or contacting them when the algorithm shows a UTI risk for an individual. The labelled data is validated based on clinical testing and is recorded by our clinical monitoring team. This data is used to train and test the binary classifiers.

\section{Evaluation Results} \label{sec result} 
The evaluation is performed on a dataset collected from 110 homes of PwD. The dataset contained a large set of unlabelled data representing 3,864 days and labelled data representing $60$ days. 
A meta-analysis on history of uncomplicated UTIs shows that most patients improve or become symptom free spontaneously, in the first 9 days of a UTI \cite{hoffmann2020natural}. The bacteriological cure and symptomatic relief usually occurs within 3 days from the initiation of antibiotic treatment \cite{christiaens2002randomised}. We selected the day that a UTI was diagnosed along with three consecutive days as UTI, and the day that a UTI was misdiagnosed along with two consecutive days as non-UTI. We then used the augmented labelled data to train the classifiers and utilised a $k$-fold cross-validation technique (with $k=5$) to train and validate our proposed model (4 folds as train set and 1 fold as test set in each iteration).

To compare the proposed semi-supervised approach with existing methods, we performed three sets of experiments: (i) we used conventional supervised classifiers on the small set of labelled data, (ii) applied the proposed semi-supervised method on the combination of unlabelled and labelled data with several alternative algorithms, (iii) validated the model on a different cohort to evaluate the ability of generalisation of the model. For both scenarios, we evaluated the effect of physiological parameters in detecting UTIs by adding two physiological markers (body temperature and pulse) as extra nodes to the model. For all the methods, we normalised the data using lagrangian normalisation \cite{rezvani2019new}. Shown in Eq.~(\ref{eq:lagrangian}), $x_i$ is the $i_{th}$ feature of the data, $\lambda_i$ is the stationary point corresponding to $i_{th}$ feature. We use the unlabelled data to get the \textbf{$\lambda$} by the first equation, and then get the normalised data $\hat{\textbf{x}}$ uses the second equation and the \textbf{$\lambda$}. The evaluation results are presented in two sections; one using only the environmental data and another with a combination of environmental and physiological data.
\begin{equation}
\label{eq:lagrangian}
 \textbf{x}^2_i - 4\lambda_i^2 = 0, \quad \hat{\textbf{x}} = \frac{\textbf{x}_i}{2\lambda}
\end{equation}

\subsection{Conventional Supervised Classifiers}
In the first experiment, the hourly SFPs were grouped in six consecutive hours to generate distinctive views of participants' activities for the morning (06:00 AM to 12:00PM), afternoon (12:00PM to 6:00PM), evening (6:00PM to 0:00AM) and night (0:00AM to 257 6:00AM). Each day is represented as a $4xN$ matrix, where $N$ is the number of nodes and 4 refers to temporal windows. In addition to these four dimensions, other statistical parameters are calculated and considered as additional features. This included daily statistics of recordings from each node (e.g. maximum, minimum, mean and median) and also the numerical difference of two consecutive temporal windows. Therefore, each day is ultimately represented as an $11xN$ input matrix and considered as an individual sample for the supervised classifier algorithm. We tested our model with other well-known methods including Gaussian Na\"ive Bayes (GB), Logistic Regression (LR), Support Vector Machine (SVM) with RBF and Polynomial kernels, Decision Tree (DT) and K-nearest Neighbour (KNN) algorithms to classify the data and provided numerical assessments. The results are shown in Table~\ref{tab1}. 

\begin{figure*}[ht!]
\begin{subfigure}[b]{0.31\textwidth}
    \centering
    \includegraphics[width=\linewidth]{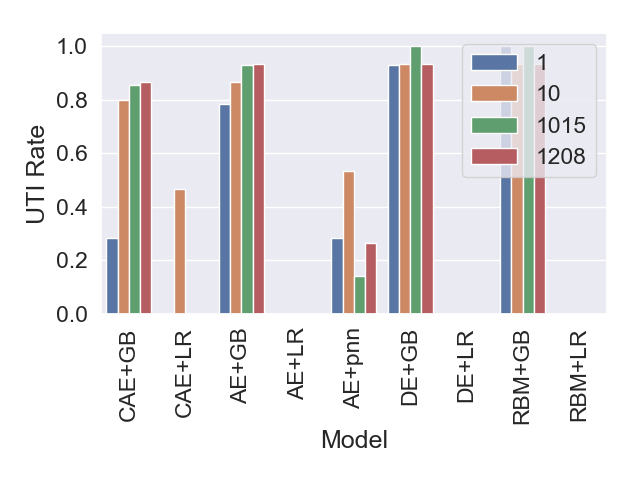}
    \caption{The rate of alerts generated by the models during the testing phase.}
    \label{fig:unlabel_alerts}
\end{subfigure}
\hfill
\begin{subfigure}[b]{0.31\textwidth}
    \centering
    \includegraphics[width=\linewidth]{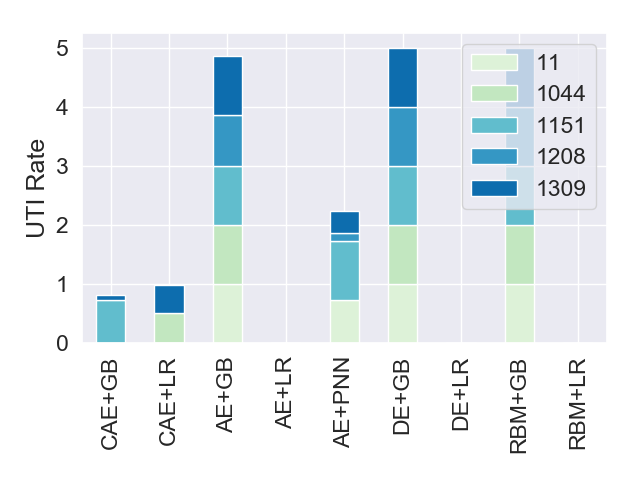}
    \caption{The rate of alerts generated by the models. The patients are validated as negative UTI.}
    \label{fig:negative_alerts}
\end{subfigure}
\hfill
\begin{subfigure}[b]{0.31\textwidth}
    \centering
    \includegraphics[width=\linewidth]{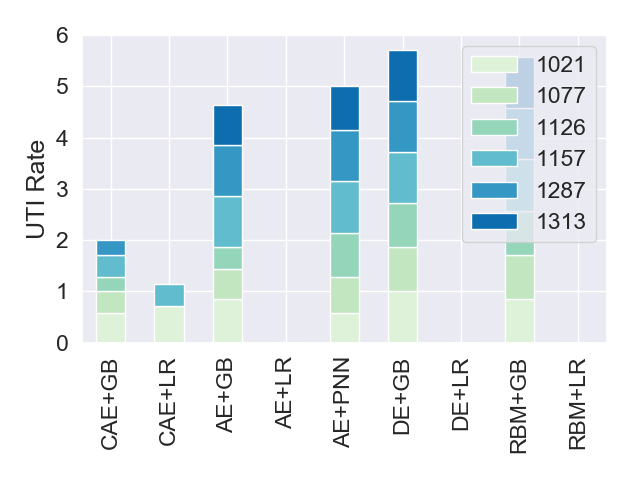}
    \caption{The rate of alerts generated by the models. The patients are validated as positive UTI.}
    \label{fig:positive_alerts}
\end{subfigure}
\caption{Testing the model on a new cohort and with the data that was not seen by the trained model before.T he AE+LR model did not detect any positive UTI cases. We have shown the model in the results but the column shows zero.}
\end{figure*}

To enhance the performance of GB and LR classifiers, we also applied two different feature selection techniques: Sequential Backward Selection (SBS)~\cite{aha1996comparative} and Recursive Feature Elimination with Cross-Validation (RFECV). The SBS and RFECV are used to reduce the feature space of the data from $n$ into $d$ dimension where $d<n$. SBS keeps removing the feature from the subset based on the performance of the classifier until the subset contains $d$ features~\cite{aha1996comparative} where $d$ is a pre-defined parameter. Given that the full feature space is $\mathbf{F} = \{f_1,f_2,\dots,f_n\}$, the subset of feature space at the initial state is defined as $\mathbf{F}^* = \mathbf{F}$. At each iteration, SBS removes feature $f_i$ from $\mathbf{F}^*$, where $f_i$ has the least contribution to the overall performance of the classifier. This iteration continues until the final feature space $\mathbf{F}^*$ has $d$ remaining features. The process of RFECV is very similar to SBS, except that a cross-validation approach is used to improve the performance. The results are shown in Table~\ref{tab1}

\setlength{\tabcolsep}{2pt}
\begin{table*}[t]
\captionof{table}{Performance of conventional supervised classifiers with and without feature selection algorithms, using only the environmental data (left) vs. the combination of environmental and physiological data (right).}
\centering
\begin{tabular}
{|>{\centering\arraybackslash}p{4cm} 
|>{\centering\arraybackslash}p{1.5cm}
|>{\centering\arraybackslash}p{1.5cm}
|>{\centering\arraybackslash}p{1.5cm}
|>{\centering\arraybackslash}p{1.5cm}
||>{\centering\arraybackslash}p{1.5cm}
|>{\centering\arraybackslash}p{1.5cm}
|>{\centering\arraybackslash}p{1.5cm}
|>{\centering\arraybackslash}p{1.5cm}|}
\hline 
& \multicolumn{4}{|c||}{Evaluation} & \multicolumn{4}{|c|}{Evaluation}  \\
Technique & \multicolumn{4}{|c||}{using Environmental Data} & \multicolumn{4}{|c|}{using Environmental and Physiological Data} \\ 
 & Precision & Recall & F1-Score & Accuracy & Precision & Recall & F1-Score & Accuracy\\ \hline
 GB              & 0.60  &  0.58  &  0.58  &  0.61 & 0.60 & 0.59 & 0.58 & 0.62\\ \hline 
 LR               & 0.71  &  0.72  &  0.69  &  0.77 & 0.70 & 0.71 & 0.68 & 0.75\\ \hline 
 SVM (Polynomial kernel)   & 0.30  &  0.5  &  0.37  &  0.60&  0.30 & 0.50 & 0.37 & 0.60\\ \hline
 SVM (RBF kernel)         & 0.30  &  0.5  &  0.37  &  0.60 & 0.30 & 0.50 & 0.37 & 0.60\\ \hline
 DT               & 0.66  &  0.60  &  0.59  &  0.66 & 0.66 & 0.60 & 0.58 & 0.63\\ \hline
 KNN              & 0.30  &  0.5  &  0.37  &  0.60 & 0.30 & 0.50 & 0.38 & 0.60\\ \hline 
 GB + SBS        & 0.76 & 0.77 & 0.75 & 0.81 & 0.77 & 0.70 & 0.68 & 0.77\\ \hline
 LR + SBS         & 0.75 & 0.67 & 0.66 & 0.72 & 0.70 & 0.63 & 0.62 & 0.69\\ \hline
 LR + REFCV       & 0.81 & 0.74 & 0.74 & 0.77 & 0.78 & 0.72 & 0.71 & 0.77\\ \hline 
\end{tabular}
\vspace{7pt}  \label{tab1}
\end{table*}
\setlength{\tabcolsep}{2pt}
\begin{table*}[t]
\captionof{table}{Performance of the classifiers within the proposed semi-supervised method, using only the environmental data (left) vs. the combination of environmental and physiological data (right).}
\centering
\begin{tabular}
{|>{\centering\arraybackslash}p{4cm} 
|>{\centering\arraybackslash}p{1.5cm}
|>{\centering\arraybackslash}p{1.5cm}
|>{\centering\arraybackslash}p{1.5cm}
|>{\centering\arraybackslash}p{1.5cm}
||>{\centering\arraybackslash}p{1.5cm}
|>{\centering\arraybackslash}p{1.5cm}
|>{\centering\arraybackslash}p{1.5cm}
|>{\centering\arraybackslash}p{1.5cm}|}
\hline 
& \multicolumn{4}{|c||}{Evaluation} & \multicolumn{4}{|c|}{Evaluation}  \\
Technique & \multicolumn{4}{|c||}{using Environmental Data} & \multicolumn{4}{|c|}{using Environmental and Physiological Data} \\ 
 & Precision & Recall & F1-Score & Accuracy & Precision & Recall & F1-Score & Accuracy\\ \hline
 CAE + GB      & 0.69 & 0.67 & 0.67 & 0.70 &  0.73 & 0.68 & 0.68 & 0.71  \\ \hline
 AE + GB       & 0.72 & 0.71 & 0.70 & 0.73 &  0.73 & 0.72 & 0.71 & 0.71  \\ \hline
 DE + GB       & 0.72 & 0.63 & 0.62 & 0.67 &  0.77 & 0.72 & 0.70 & 0.73  \\ \hline
 RBM + GB      & 0.61 & 0.63 & 0.59 & 0.65 &  0.61 & 0.63 & 0.59 & 0.65 \\ \hline
 CAE + LR      & 0.81 & 0.74 & 0.74 & 0.78 & 0.81 & 0.74 & 0.73 & 0.77  \\ \hline
 AE + LR       & 0.72 & 0.65 & 0.61 & 0.71 & 0.78 & 0.67 & 0.65 & 0.73  \\ \hline
 DE + LR        & 0.79 & 0.75 & 0.72 & 0.80 & 0.74 & 0.64 & 0.61 & 0.71  \\ \hline
 RBM + LR       & 0.53 & 0.57 & 0.52 & 0.63 &  0.54 & 0.59 & 0.54 & 0.65 \\ \hline
\textbf{AE + PNN}       &0.83 & 0.82 & 0.81 & 0.83 &  0.86 & 0.85 & 0.83 & 0.85  \\ \hline
\end{tabular}
\vspace{7pt}  \label{tab2}
\end{table*}

\subsection{Semi-supervised learning}
In the second experiment, we applied our semi-supervised model to the combination of unlabelled and unlabelled data. We used a Convolutional auto-encoder (AE), Deep-encoder (DE), and Restricted Boltzmann Machine (RBM) to represent the unlabelled data~\cite{bengio2013representation}. The unlabelled data is aggregated hourly and each day is represented as a $24xN$ matrix, where $N$ is the number of sensory sources. The input of the semi-supervised method is constructed as a $3864x24xN$ matrix which is fed to the unsupervised learning techniques to extract latent features. We searched for the most optimised hyper-parameters for all the methods. In the CAE, the encoder contains two convolutional layers which contain 16, 38 $3\times 3$ filters; the decoder contains two convolutional layers which contain 8, 1 $3\times 3$ filters. The latent dimension is 7296. In the AE, the encoder and decoder contain one fully-connected layer, and the latent dimension is 171. In the DE, the encoder contains three fully-connected layers with 128, 64, 86 neurons; the decoder contains three fully-connected layers with 64, 128, 192 neurons respectively. The latent dimension is 20. The generated model in the unsupervised step is then applied to the daily aggregated labelled data to provide the required features for the supervised classifiers. In the PNN, we use the same architecture as the AE. We use the Adadelta as optimiser and binary cross-entropy to train the AE, use Adam as optimiser and mean squared error to train the PNN. The results are shown in Table~\ref{tab2}. To be consistent across all the experiments, we train all the models in 5000 iterations and set the learning rate to 0.01. 

\begin{figure}[h]
    \includegraphics[width=0.8\linewidth]{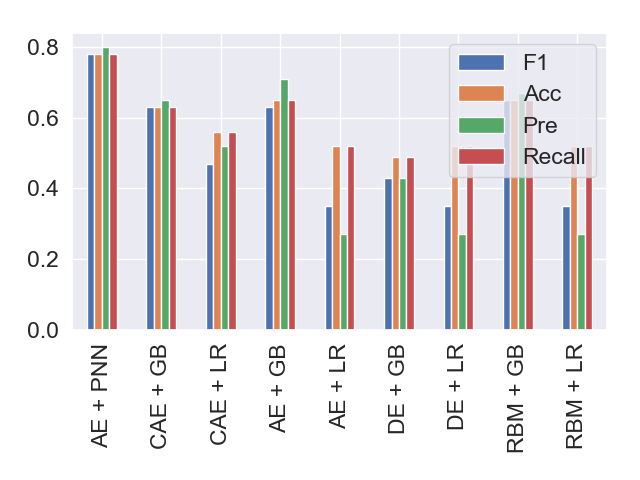}
    \caption{Five fold cross-validation results on the new cohort.}
    \label{fig:dri_results}
\end{figure}
\begin{figure}[h]
    \includegraphics[width=0.8\linewidth]{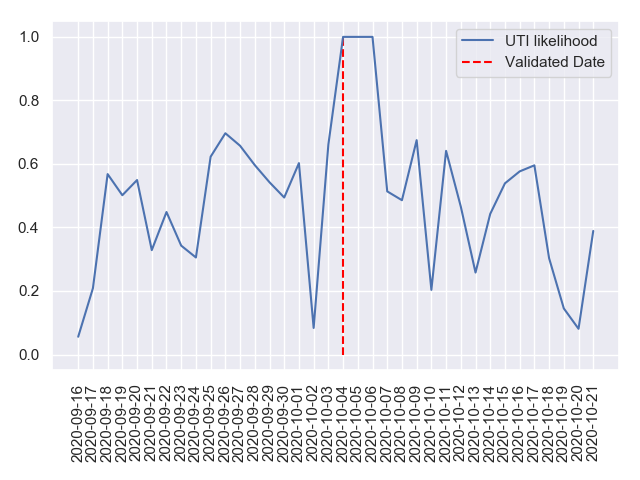}
    \caption{UTI likelihood changes of a patient. }
    \label{fig:uti_1313}
\end{figure}

\subsection{Generalisation}
One of the inevitable issues of deploying the model in the real-world is generalisation. This issue becomes more pertinent when the training set is not sufficiently large. We further analyse the models' ability to generalise in the real-world setting especially when transferring the model to a completely new cohort. To do this, we have used a new set of data that is collected in our study (which were not seen by the algorithms before) and test the algorithms. The new data set collected from 68 patients (32 males and 36 females) with a mean of age 81 and a standard deviation 13.7. Fig. \ref{fig:unlabel_alerts} shows a snapshot of the alerts generated by the models during this evaluation. The models are tested on 13,090 days of patient data and produce the predictions. Some of the models overfit the training set leading to relatively high UTI rates (e.g. CAE + GB), or result in relatively low UTI rates (e.g. DE + LR). The AE + PNN  method UTI risk detection reduces the numbers of false alerts. During the deployment, the clinical monitoring team validated the generated alerts. Shown in Fig. \ref{fig:negative_alerts} and \ref{fig:positive_alerts} are sample cases validated as no-UTI and UTI respectively. 
The proposed model produces a relatively small number of alerts on the no-UTI cases and is sufficiently sensitive to the UTI cases. After collecting the labelled data in the new cohort, we have tested the performance of the model again (see Fig. \ref{fig:dri_results}). The proposed model outperforms the baseline models on the new cohort as well.

\section{Discussion} 
Our proposed semi-supervised learning trains an auto-encoder and PNN simultaneously. As shown in Table~\ref{tab1} and \ref{tab2}, the use of semi-supervised approach improved the performances of the baseline classifiers. The semi-supervised learning methods significantly enhanced the performance of other classifiers, such as GB and LR. It is while our proposed model combining the fully connected AE and PNN outperformed all other methods and provided the highest reliability which achieves 83\% F1-accuracy. The recall in the tables represents the average of sensitivity and specificity. The average values show that the models' perform well when predicting both positive and negative UTI cases. The results show that the proposed model performs the best and achieves 85\% recall. These evaluation results show that the proposed model performs better on detecting the UTI events than others.

Interestingly, adding the physiological markers to the conventional classification approaches reduced the accuracy of some of them. This could be due to the fact that clinical devices required user interactions to collect the measurements and there were less frequent daily samples when compared with the environmental observations, which were recorded continuously via passive sensors. The recording rate (i.e. frequency) of the environmental data could raise as high as one sample per second depending on the activity level within the home environment, while the physiological data was generally observed once or twice a day. In some cases, the users did not take the measurements regularly, leading to missing values in the data. Such heterogeneity in recording physiological parameters affected the learning process, resulting in lower F1-score in using both physiological and environmental components.

The proposed semi-supervised model outperforms even the modified classifiers (i.e. classifiers combined with the feature selection techniques). One of the benefits of the semi-supervised method is that it can be applied to the hourly aggregated data without any additional statistical features. Furthermore, this approach learns based on unlabelled datasets collected from uncontrolled environments; making it more flexible and generalisable for real-world applications, especially in cases where only a small set of labelled data is available and/or when the event of interest is a rare phenomenon. This is particularly important for healthcare applications where collecting labelled data is either a time/resource consuming task or very challenging due to the low prevalence of the event of interest. For instance, the UTI prevalence for women and men over the age of 85 is 13\% and 8\%~\cite{caljouw2011predictive}. This means UTIs occur sporadically within a relatively large dataset. Hence, our semi-supervised learning model utilises the existing (but unlabelled) data to represent the data and then uses a probabilistic model to train the feature space with the given examples (i.e. the few examples of UTI). The proposed model can continuously learn from the new samples without retraining.

\begin{figure*}
\begin{subfigure}[b]{0.24\textwidth}
    \centering
    \includegraphics[width=\linewidth]{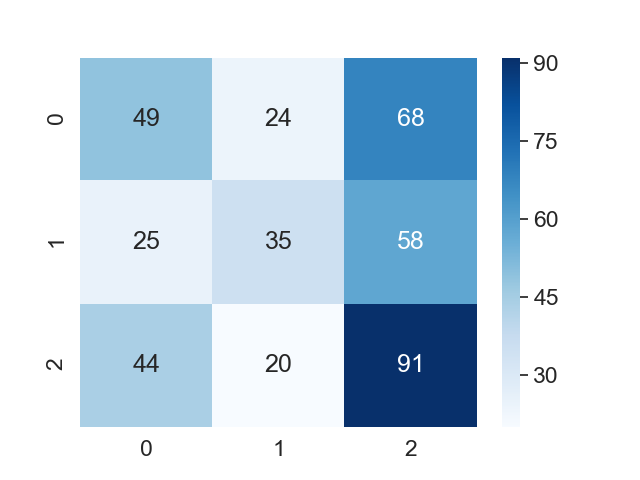}
    \caption{Confusion matrix of the proposed model. Each phase contains data for 10 days.}
    \label{fig:cf_pnn}
\end{subfigure}
\begin{subfigure}[b]{0.24\textwidth}
    \centering
    \includegraphics[width=\linewidth]{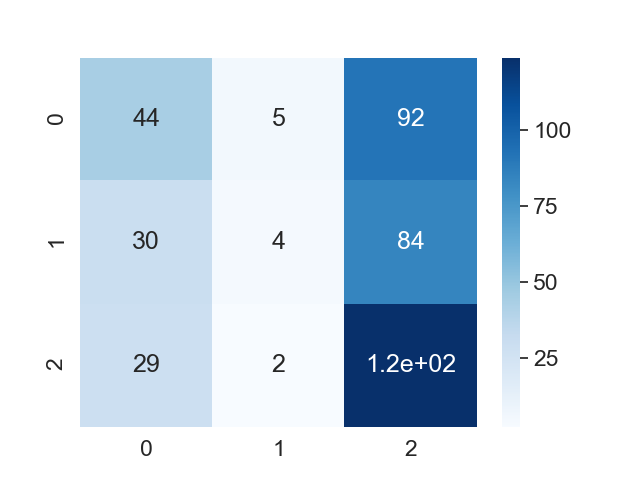}
    \caption{Confusion matrix of the baseline model (CAE + LR). Each phase contains data for 10 days.}
    \label{fig:cf_lr}
\end{subfigure}
\begin{subfigure}[b]{0.24\textwidth}
    \centering
    \includegraphics[width=\linewidth]{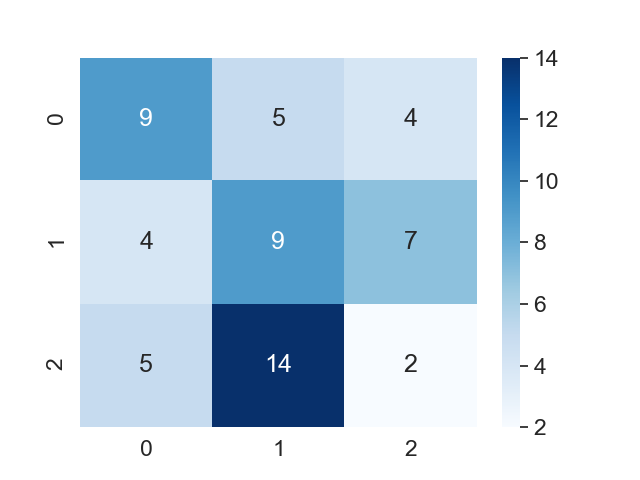}
    \caption{Confusion matrix of the proposed model. Each phase contains data for 5 days.}
    \label{fig:5_cf_pnn}
\end{subfigure}
\begin{subfigure}[b]{0.24\textwidth}
    \centering
    \includegraphics[width=\linewidth]{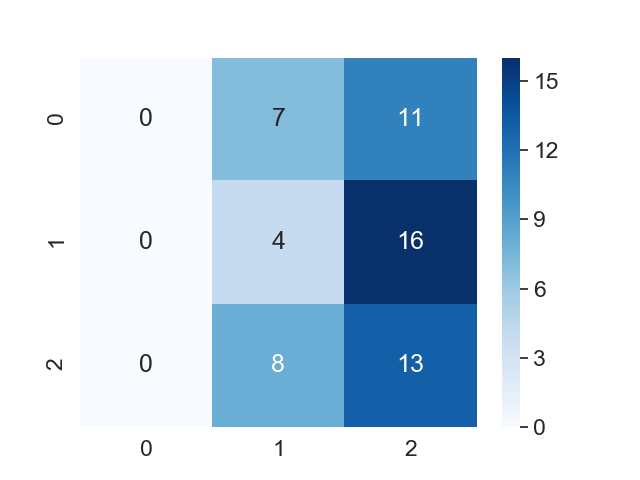}
    \caption{Confusion matrix of the baseline model (CAE + LR). Each phase contains data for 5 days.}
    \label{fig:5_cf_lr}
\end{subfigure}
\caption{Confusion matrices for the evaluations}
\end{figure*}

The model is also able to distinguish between different stages of the infection. We defined three UTI phases: pre-UTI, UTI and post-UTI, each lasting 10 days with the validated UTI diagnosis date in the middle of this 30-day period. We then compared the likelihood of UTI across these three periods and showed that it significantly increases with the proximity to the validate diagnosis date (Fig. \ref{fig:uti_1313} shows the likelihood of diagnosed as UTI). We then tested whether our proposed model is able to distinguish between the three UTI phases, and subsequently generate an early alarm of the UTI. We also conducted the same analysis using CAE + LR, which performed best among the other baseline models. Our proposed model achieved 40.88\% F1 accuracy in distinguishing between the three phases, compared to 33.78\% f1 accuracy using the CAE + LR model. Fig. \ref{fig:cf_pnn} and \ref{fig:cf_lr} show the confusion matrix, where 0, 1 and 2 represent the pre-UTI, UTI and post-UTI phases, respectively. The confusion matrix demonstrates that the CAE + LR is more likely to overfit the samples. We have also assumed each of the phase last 5 days, the proposed model achieved 31.49\% F1 accuracy compared to 29.18\% F1 accuracy using the CAE + LR model (see Fig. \ref{fig:5_cf_pnn} and \ref{fig:5_cf_lr}).


\section{Translating the solution into clinical practice}
The 2020 report of the Lancet Commission on dementia prevention, treatment, and care stresses the importance of individualised interventions to address complex medical problems and multimorbidity in dementia, which lead to unnecessary hospital admissions, faster functional decline, and worse quality of life \cite{livingston2020dementia}. Infections have been highlighted as one of the areas for priority development to advance dementia care \cite{pickett2018roadmap}. Our proposed model directly addresses these priorities in dementia care and intervention by enabling early detection of urinary tract infections in remote healthcare monitoring setting, providing an opportunity for delivering more personalised, predictive and preventative healthcare.

When applied to real-world clinical dataset in our clinical study the proposed algorithm provided the F1-score of 83\% in detecting UTI from the physiological and environmental sensor data. All UTI predictions have been verified by a clinical monitoring team who contact the patient or carer when a UTI alert is first generated to evaluate the symptoms. The team then arranges a home visit to perform a urine dipstick test. If the urinary analysis is suggestive of infection (positive nitrates or leukocytes) clinical monitoring team advises the person with dementia or carer to visit the GP the same day to obtain a prescription for antibiotics. Monitoring Team also informs the GP of the test results and requests for antibiotics to be prescribed.

The impact of our in-home monitoring technologies and the embedded machine learning models on clinical outcomes including hospitalisation, institutionalisation and mortality rates, as well as patients’ quality of life is part of an ongoing study. Nevertheless, the current work demonstrates the effectiveness of the proposed algorithm and its translation into real-life clinical interventions.

One potential criticism of our UTI intervention algorithm could be the possibility of antibiotic over-prescribing contributing to the spread of antibiotic resistance. However, recent evidence demonstrates that in elderly patients with a diagnosis of UTI in primary care, no antibiotics and delayed antibiotics are associated with a significant increase in bloodstream infection and all-cause mortality compared with immediate treatment \cite{pickett2018roadmap}. Therefore, early prescription of antibiotics for this vulnerable group of older adults is advised in view of their increased susceptibility to sepsis after UTI and despite a growing pressure to reduce inappropriate antibiotic use.

\section{Conclusions}
Using new forms of data collected by network-enabled devices and combining environmental and physiological markers with machine learning models provide remote healthcare monitoring systems which can profoundly improve the detection of the risk of sporadic healthcare events. Remote healthcare monitoring systems allow access to new forms of data through the use of  network-enabled devices. These environmental and physiological markers, combined with machine learning models, have the potential to profoundly change the detection of sporadic healthcare events in patients living with dementia. This work discusses a novel approach for detecting the risk of Urinary Tract Infection (UTI) which is the common bacterial infection in an elderly population and one of the most frequent reasons for avoidable hospital admissions, associated with a number of negative outcomes in dementia population. The diagnosis of UTI in elderly patients with dementia remains problematic due to presence of a range of nonspecific (or atypical) symptoms, high prevalence of asymptomatic bacteriuria and reduced or delayed help-seeking behaviour. In addition, current UTI diagnostic tests  such as the urinalysis or urine culture test require advance notice of the risk prior to recommending the test. Thus, interpreting the probability of UTI based on sensory data allows for early risk recognition, leading to faster and greater accuracy in diagnosis.

In this paper, we propose a semi-supervised approach which uses a combination of labelled (small set) and unlabelled (large set) data to detect the risk of UTI in people with dementia. We then combine the semi-supervised approach with the probabilistic neural networks to enhance the performance of the prediction model. The proposed method has been evaluated in a real-world application in dementia cohort and  was found to have very good diagnostic performance with an F1-score of 83\% which outperformed the other modified supervised classifiers. 

\section*{Acknowledgements}
This research is funded by Medical Research Council (MRC), Alzheimer's Society and Alzheimer's Research UK and supported by the UK Dementia Research Institute. 
\bibliographystyle{IEEEtran}
\bibliography{refs}

\begin{thebibliography}{10}
\providecommand{\url}[1]{#1}
\csname url@samestyle\endcsname
\providecommand{\newblock}{\relax}
\providecommand{\bibinfo}[2]{#2}
\providecommand{\BIBentrySTDinterwordspacing}{\spaceskip=0pt\relax}
\providecommand{\BIBentryALTinterwordstretchfactor}{4}
\providecommand{\BIBentryALTinterwordspacing}{\spaceskip=\fontdimen2\font plus
\BIBentryALTinterwordstretchfactor\fontdimen3\font minus
  \fontdimen4\font\relax}
\providecommand{\BIBforeignlanguage}[2]{{%
\expandafter\ifx\csname l@#1\endcsname\relax
\typeout{** WARNING: IEEEtran.bst: No hyphenation pattern has been}%
\typeout{** loaded for the language `#1'. Using the pattern for}%
\typeout{** the default language instead.}%
\else
\language=\csname l@#1\endcsname
\fi
#2}}
\providecommand{\BIBdecl}{\relax}
\BIBdecl

\bibitem{linhares2013frequency}
I.~Linhares, T.~Raposo, A.~Rodrigues, and A.~Almeida, ``Frequency and
  antimicrobial resistance patterns of bacteria implicated in community urinary
  tract infections: a ten-year surveillance study (2000--2009),'' \emph{BMC
  infectious diseases}, vol.~13, no.~1, p.~19, 2013.

\bibitem{rao2016outcomes}
A.~Rao, A.~Suliman, S.~Vuik, P.~Aylin, and A.~Darzi, ``Outcomes of dementia:
  systematic review and meta-analysis of hospital administrative database
  studies,'' \emph{Archives of gerontology and geriatrics}, vol.~66, pp.
  198--204, 2016.

\bibitem{masajtis2017new}
A.~Masajtis-Zagajewska and M.~Nowicki, ``New markers of urinary tract
  infection,'' \emph{Clinica Chimica Acta}, vol. 471, pp. 286--291, 2017.

\bibitem{sampson2009dementia}
E.~L. Sampson, M.~R. Blanchard, L.~Jones, A.~Tookman, and M.~King, ``Dementia
  in the acute hospital: prospective cohort study of prevalence and
  mortality,'' \emph{The British Journal of Psychiatry}, vol. 195, no.~1, pp.
  61--66, 2009.

\bibitem{burnham2018urinary}
P.~Burnham, D.~Dadhania, M.~Heyang, F.~Chen, L.~F. Westblade, M.~Suthanthiran,
  J.~R. Lee, and I.~De~Vlaminck, ``Urinary cell-free dna is a versatile analyte
  for monitoring infections of the urinary tract,'' \emph{Nature
  communications}, vol.~9, no.~1, pp. 1--10, 2018.

\bibitem{gavazzi2013diagnostic}
G.~Gavazzi, E.~Delerce, E.~Cambau, P.~Fran{\c{c}}ois, B.~Corroyer,
  B.~de~Wazi{\`e}res, B.~Foug{\`e}re, M.~Paccalin, and J.~Gaillat, ``Diagnostic
  criteria for urinary tract infection in hospitalized elderly patients over 75
  years of age: a multicenter cross-sectional study,'' \emph{M{\'e}decine et
  maladies infectieuses}, vol.~43, no.~5, pp. 189--194, 2013.

\bibitem{nicolle2003asymptomatic}
L.~E. Nicolle, ``Asymptomatic bacteriuria: when to screen and when to treat,''
  \emph{Infectious Disease Clinics}, vol.~17, no.~2, pp. 367--394, 2003.

\bibitem{marques2012epidemiological}
L.~P.~J. Marques, J.~T. Flores, O.~d. O.~B. Junior, G.~B. Rodrigues,
  C.~de~Medeiros~Mour{\~a}o, and R.~M.~P. Moreira, ``Epidemiological and
  clinical aspects of urinary tract infection in community-dwelling elderly
  women,'' \emph{The Brazilian Journal of infectious diseases}, vol.~16, no.~5,
  pp. 436--441, 2012.

\bibitem{chu2018diagnosis}
C.~M. Chu and et~al., ``Diagnosis and treatment of urinary tract infections
  across age groups,'' \emph{Ame. J. Obst. \& Gyne}, 2018.

\bibitem{peach2016risk}
B.~C. Peach, G.~J. Garvan, C.~S. Garvan, and J.~P. Cimiotti, ``Risk factors for
  urosepsis in older adults: a systematic review,'' \emph{Gerontology and
  geriatric medicine}, vol.~2, p. 2333721416638980, 2016.

\bibitem{tal2005profile}
S.~Tal, V.~Guller, S.~Levi, R.~Bardenstein, D.~Berger, I.~Gurevich, and
  A.~Gurevich, ``Profile and prognosis of febrile elderly patients with
  bacteremic urinary tract infection,'' \emph{Journal of Infection}, vol.~50,
  no.~4, pp. 296--305, 2005.

\bibitem{gharbi2019antibiotic}
M.~Gharbi, J.~H. Drysdale, H.~Lishman, R.~Goudie, M.~Molokhia, A.~P. Johnson,
  A.~H. Holmes, and P.~Aylin, ``Antibiotic management of urinary tract
  infection in elderly patients in primary care and its association with
  bloodstream infections and all cause mortality: population based cohort
  study,'' \emph{bmj}, vol. 364, 2019.

\bibitem{toot2013causes}
S.~Toot, M.~Devine, A.~Akporobaro, and M.~Orrell, ``Causes of hospital
  admission for people with dementia: a systematic review and meta-analysis,''
  \emph{Journal of the American Medical Directors Association}, vol.~14, no.~7,
  pp. 463--470, 2013.

\bibitem{little2006developing}
P.~Little, S.~Turner, K.~Rumsby, G.~Warner, M.~Moore, J.~A. Lowes, H.~Smith,
  C.~Hawke, and M.~Mullee, ``Developing clinical rules to predict urinary tract
  infection in primary care settings: sensitivity and specificity of near
  patient tests (dipsticks) and clinical scores,'' \emph{British journal of
  general practice}, vol.~56, no. 529, pp. 606--612, 2006.

\bibitem{mcisaac2007validation}
W.~J. McIsaac, R.~Moineddin, and S.~Ross, ``Validation of a decision aid to
  assist physicians in reducing unnecessary antibiotic drug use for acute
  cystitis,'' \emph{Archives of internal medicine}, vol. 167, no.~20, pp.
  2201--2206, 2007.

\bibitem{heckerling2007predictors}
P.~S. Heckerling, G.~J. Canaris, S.~D. Flach, T.~G. Tape, R.~S. Wigton, and
  B.~S. Gerber, ``Predictors of urinary tract infection based on artificial
  neural networks and genetic algorithms,'' \emph{International Journal of
  Medical Informatics}, vol.~76, no.~4, pp. 289--296, 2007.

\bibitem{papageorgiou2012fuzzy}
E.~I. Papageorgiou, ``Fuzzy cognitive map software tool for treatment
  management of uncomplicated urinary tract infection,'' \emph{Computer methods
  and programs in biomedicine}, vol. 105, no.~3, pp. 233--245, 2012.

\bibitem{majumder2017smart}
S.~Majumder, E.~Aghayi, M.~Noferesti, H.~Memarzadeh-Tehran, T.~Mondal, Z.~Pang,
  and M.~J. Deen, ``Smart homes for elderly healthcare—recent advances and
  research challenges,'' \emph{Sensors}, vol.~17, no.~11, p. 2496, 2017.

\bibitem{turjamaa2019smart}
R.~Turjamaa, A.~Pehkonen, and M.~Kangasniemi, ``How smart homes are used to
  support older people: An integrative review,'' \emph{International Journal of
  Older People Nursing}, vol.~14, no.~4, p. e12260, 2019.

\bibitem{peetoom2015literature}
K.~K. Peetoom, M.~A. Lexis, M.~Joore, C.~D. Dirksen, and L.~P. De~Witte,
  ``Literature review on monitoring technologies and their outcomes in
  independently living elderly people,'' \emph{Disability and Rehabilitation:
  Assistive Technology}, vol.~10, no.~4, pp. 271--294, 2015.

\bibitem{schwickert2013farseeing}
L.~Schwickert, C.~Becker, U.~Lindemann, C.~Mar{\'e}chal, A.~Bourke, L.~Chiari,
  J.~Helbostad, W.~Zijlstra, K.~Aminian, C.~Todd \emph{et~al.}, ``Farseeing
  consortium and the farseeing meta database consensus group,'' \emph{Fall
  detection with body-worn sensors: a systematic review. Z Gerontol Geriatr},
  vol.~46, pp. 706--719, 2013.

\bibitem{lazarou2016novel}
I.~Lazarou, A.~Karakostas, T.~G. Stavropoulos, T.~Tsompanidis, G.~Meditskos,
  I.~Kompatsiaris, and M.~Tsolaki, ``A novel and intelligent home monitoring
  system for care support of elders with cognitive impairment,'' \emph{Journal
  of Alzheimer's Disease}, vol.~54, no.~4, pp. 1561--1591, 2016.

\bibitem{bankole2012validation}
A.~Bankole, M.~Anderson, T.~Smith-Jackson, A.~Knight, K.~Oh, J.~Brantley,
  A.~Barth, and J.~Lach, ``Validation of noninvasive body sensor network
  technology in the detection of agitation in dementia,'' \emph{American
  Journal of Alzheimer's Disease \& Other Dementias}, vol.~27, no.~5, pp.
  346--354, 2012.

\bibitem{fleiner2016sensor}
T.~Fleiner, P.~Haussermann, S.~Mellone, and W.~Zijlstra, ``Sensor-based
  assessment of mobility-related behavior in dementia: feasibility and
  relevance in a hospital context,'' \emph{International Psychogeriatrics},
  vol.~28, no.~10, p. 1687, 2016.

\bibitem{rantz2011using}
M.~J. Rantz, M.~Skubic, R.~J. Koopman, L.~Phillips, G.~L. Alexander, S.~J.
  Miller, and R.~D. Guevara, ``Using sensor networks to detect urinary tract
  infections in older adults,'' \emph{Proceedings of IEEE International
  Conference on e-Health Networking Applications and Services (Healthcom)}, pp.
  142--149, 2011.

\bibitem{enshaeifar2019machine}
S.~Enshaeifar, A.~Zoha, S.~Skillman, A.~Markides, S.~T. Acton, T.~Elsaleh,
  M.~Kenny, H.~Rostill, R.~Nilforooshan, and P.~Barnaghi, ``Machine learning
  methods for detecting urinary tract infection and analysing daily living
  activities in people with dementia,'' \emph{PloS one}, vol.~14, no.~1, p.
  e0209909, 2019.

\bibitem{enshaeifar2018health}
S.~Enshaeifar, A.~Zoha, A.~Markides, S.~Skillman, S.~T. Acton, T.~Elsaleh,
  M.~Hassanpour, A.~Ahrabian, M.~Kenny, S.~Klein \emph{et~al.}, ``Health
  management and pattern analysis of daily living activities of people with
  dementia using in-home sensors and machine learning techniques,'' \emph{PloS
  one}, vol.~13, no.~5, p. e0195605, 2018.

\bibitem{enshaeifar2018UTI}
S.~Enshaeifar, A.~Zoha, S.~Skillman, A.~Markides, S.~T. Acton, T.~Elsaleh,
  M.~Kenny, H.~Rostill, R.~Nilforooshan, and P.~Barnaghi, ``Machine learning
  methods for detecting urinary tract infection and analysing daily living
  activities in people with dementia,'' \emph{PloS one (in Press)}, 2018.

\bibitem{doukas2012bringing}
C.~Doukas and I.~Maglogiannis, ``Bringing {IoT} and cloud computing towards
  pervasive healthcare,'' \emph{Proceedings of IEEE International Conference on
  Innovative Mobile and Internet Services in Ubiquitous Computing (IMIS)}, pp.
  922--926, 2012.

\bibitem{yang2014health}
G.~Yang, L.~Xie, M.~M{\"a}ntysalo, X.~Zhou, Z.~Pang, L.~Da~Xu, S.~Kao-Walter,
  Q.~Chen, and L.-R. Zheng, ``A health-{IoT} platform based on the integration
  of intelligent packaging, unobtrusive bio-sensor, and intelligent medicine
  box,'' \emph{IEEE Transactions on Industrial Informatics}, vol.~10, no.~4,
  pp. 2180--2191, 2014.

\bibitem{catarinucci2015iot}
L.~Catarinucci, D.~De~Donno, L.~Mainetti, L.~Palano, L.~Patrono, M.~L.
  Stefanizzi, and L.~Tarricone, ``An {IoT}-aware architecture for smart
  healthcare systems,'' \emph{IEEE Internet of Things}, vol.~2, no.~6, pp.
  515--526, 2015.

\bibitem{amendola2014rfid}
S.~Amendola, R.~Lodato, S.~Manzari, C.~Occhiuzzi, and G.~Marrocco, ``{RFID}
  technology for {IoT}-based personal healthcare in smart spaces,'' \emph{IEEE
  Internet of Things}, vol.~1, no.~2, pp. 144--152, 2014.

\bibitem{coates2011analysis}
A.~Coates, A.~Ng, and H.~Lee, ``An analysis of single-layer networks in
  unsupervised feature learning,'' in \emph{Proceedings of the fourteenth
  international conference on artificial intelligence and statistics}, 2011,
  pp. 215--223.

\bibitem{sheikhpour2017survey}
R.~Sheikhpour and et~al., ``A survey on semi-supervised feature selection
  methods,'' \emph{Pattern Recognition}, vol.~64, pp. 141--158, 2017.

\bibitem{zhu2006semi}
X.~Zhu, ``Semi-supervised learning literature survey,'' \emph{Computer Science,
  University of Wisconsin-Madison}, vol.~2, no.~3, p.~4, 2006.

\bibitem{prakash2014survey}
V.~J. Prakash and D.~L. Nithya, ``A survey on semi-supervised learning
  techniques,'' \emph{arXiv preprint arXiv:1402.4645}, 2014.

\bibitem{zhu2009introduction}
X.~Zhu and A.~B. Goldberg, ``Introduction to semi-supervised learning,''
  \emph{Synthesis lectures on {AI and ML}}, vol.~3, no.~1, pp. 1--130, 2009.

\bibitem{zhu2003semi}
X.~Zhu, Z.~Ghahramani, and J.~D. Lafferty, ``Semi-supervised learning using
  gaussian fields and harmonic functions,'' in \emph{Proceedings of the 20th
  International conference on Machine learning (ICML-03)}, 2003, pp. 912--919.

\bibitem{TIHM2020}
S.~Enshaeifar, P.~Barnaghi, S.~Skillman, D.~Sharp, R.~Nilforooshan, and
  H.~Rostill, ``A digital platform for remote healthcare monitoring,'' in
  \emph{Companion Proceedings of the Web Conference 2020}, ser. WWW
  ’20.\hskip 1em plus 0.5em minus 0.4em\relax New York, NY, USA: Association
  for Computing Machinery, 2020, p. 203–206.

\bibitem{schmidhuber2015deep}
J.~Schmidhuber, ``Deep learning in neural networks: An overview,'' \emph{Neural
  networks}, vol.~61, pp. 85--117, 2015.

\bibitem{le2015tutorial}
Q.~V. Le \emph{et~al.}, ``A tutorial on deep learning part 2: autoencoders,
  convolutional neural networks and recurrent neural networks,'' \emph{Google
  Brain}, pp. 1--20, 2015.

\bibitem{holden2015learning}
D.~Holden, J.~Saito, T.~Komura, and T.~Joyce, ``Learning motion manifolds with
  convolutional autoencoders,'' in \emph{SIGGRAPH Asia 2015 Technical
  Briefs}.\hskip 1em plus 0.5em minus 0.4em\relax ACM, 2015, p.~18.

\bibitem{specht1990probabilistic}
D.~F. Specht, ``Probabilistic neural networks,'' \emph{Neural networks},
  vol.~3, no.~1, pp. 109--118, 1990.

\bibitem{li2020continual}
H.~Li, P.~Barnaghi, S.~Enshaeifar, and F.~Ganz, ``Continual learning using task
  conditional neural networks,'' 2020.

\bibitem{mccloskey1989catastrophic}
M.~McCloskey and N.~J. Cohen, ``Catastrophic interference in connectionist
  networks: The sequential learning problem,'' in \emph{Psychology of learning
  and motivation}.\hskip 1em plus 0.5em minus 0.4em\relax Elsevier, 1989,
  vol.~24, pp. 109--165.

\bibitem{van2006frail}
M.~Van~Iersel, A.~Verbeek, B.~Bloem, M.~Munneke, R.~A. Esselink, and M.~O.
  Rikkert, ``Frail elderly patients with dementia go too fast,'' \emph{Journal
  of Neurology, Neurosurgery \& Psychiatry}, vol.~77, no.~7, pp. 874--876,
  2006.

\bibitem{castell2013frailty}
M.-V. Castell, M.~S{\'a}nchez, R.~Juli{\'a}n, R.~Queipo, S.~Mart{\'\i}n, and
  {\'A}.~Otero, ``Frailty prevalence and slow walking speed in persons age 65
  and older: implications for primary care,'' \emph{BMC family practice},
  vol.~14, no.~1, p.~86, 2013.

\bibitem{grande2019measuring}
G.~Grande, F.~Triolo, A.~Nuara, A.-K. Welmer, L.~Fratiglioni, and D.~L.
  Vetrano, ``Measuring gait speed to better identify prodromal dementia,''
  \emph{Experimental gerontology}, vol. 124, p. 110625, 2019.

\bibitem{shimada2018cognitive}
H.~Shimada, S.~Lee, H.~Makizako, L.-K. Chen, H.~Arai \emph{et~al.}, ``Cognitive
  frailty predicts incident dementia among community-dwelling older people,''
  \emph{Journal of clinical medicine}, vol.~7, no.~9, p. 250, 2018.

\bibitem{daly2000predicting}
E.~Daly, D.~Zaitchik, M.~Copeland, J.~Schmahmann, J.~Gunther, and M.~Albert,
  ``Predicting conversion to alzheimer disease using standardized clinical
  information,'' \emph{Archives of neurology}, vol.~57, no.~5, pp. 675--680,
  2000.

\bibitem{hoang2016effect}
T.~D. Hoang, J.~Reis, N.~Zhu, D.~R. Jacobs, L.~J. Launer, R.~A. Whitmer,
  S.~Sidney, and K.~Yaffe, ``Effect of early adult patterns of physical
  activity and television viewing on midlife cognitive function,'' \emph{JAMA
  psychiatry}, vol.~73, no.~1, pp. 73--79, 2016.

\bibitem{fancourt2019television}
D.~Fancourt and A.~Steptoe, ``Television viewing and cognitive decline in older
  age: findings from the english longitudinal study of ageing,''
  \emph{Scientific reports}, vol.~9, no.~1, pp. 1--8, 2019.

\bibitem{sibbett2017risk}
R.~A. Sibbett, T.~C. Russ, I.~J. Deary, and J.~M. Starr, ``Risk factors for
  dementia in the ninth decade of life and beyond: a study of the lothian birth
  cohort 1921,'' \emph{BMC psychiatry}, vol.~17, no.~1, p. 205, 2017.

\bibitem{gilsanz2017female}
P.~Gilsanz, E.~R. Mayeda, M.~M. Glymour, C.~P. Quesenberry, D.~M. Mungas,
  C.~DeCarli, A.~Dean, and R.~A. Whitmer, ``Female sex, early-onset
  hypertension, and risk of dementia,'' \emph{Neurology}, vol.~89, no.~18, pp.
  1886--1893, 2017.

\bibitem{agata2013challenges}
E.~D. Agata, M.~B. Loeb, and S.~L. Mitchell, ``Challenges in assessing nursing
  home residents with advanced dementia for suspected urinary tract
  infections,'' \emph{Journal of the American Geriatrics Society}, vol.~61,
  no.~1, pp. 62--66, 2013.

\bibitem{juthani2007diagnostic}
M.~Juthani-Mehta, M.~Tinetti, E.~Perrelli, V.~Towle, P.~H. Van~Ness, and
  V.~Quagliarello, ``Diagnostic accuracy of criteria for urinary tract
  infection in a cohort of nursing home residents,'' \emph{Journal of the
  American Geriatrics Society}, vol.~55, no.~7, pp. 1072--1077, 2007.

\bibitem{ijaopo2017dementia}
E.~Ijaopo, ``Dementia-related agitation: a review of non-pharmacological
  interventions and analysis of risks and benefits of pharmacotherapy,''
  \emph{Translational psychiatry}, vol.~7, no.~10, pp. e1250--e1250, 2017.

\bibitem{hoffmann2020natural}
T.~Hoffmann, R.~Peiris, C.~Del~Mar, G.~Cleo, and P.~Glasziou, ``Natural history
  of uncomplicated urinary tract infection without antibiotics: a systematic
  review,'' \emph{British Journal of General Practice}, vol.~70, no. 699, pp.
  e714--e722, 2020.

\bibitem{christiaens2002randomised}
T.~Christiaens, M.~De~Meyere, G.~Verschraegen, W.~Peersman, S.~Heytens, and
  J.~De~Maeseneer, ``Randomised controlled trial of nitrofurantoin versus
  placebo in the treatment of uncomplicated urinary tract infection in adult
  women.'' \emph{British Journal of General Practice}, vol.~52, no. 482, pp.
  729--734, 2002.

\bibitem{rezvani2019new}
R.~Rezvani, P.~Barnaghi, and S.~Enshaeifar, ``A new pattern representation
  method for time-series data,'' \emph{IEEE Transactions on Knowledge and Data
  Engineering}, 2019.

\bibitem{aha1996comparative}
D.~W. Aha and R.~L. Bankert, ``A comparative evaluation of sequential feature
  selection algorithms,'' \emph{Learning from data}, 1996.

\bibitem{bengio2013representation}
Y.~Bengio, A.~Courville, and P.~Vincent, ``Representation learning: A review
  and new perspectives,'' \emph{IEEE transactions on pattern analysis and
  machine intelligence}, vol.~35, no.~8, pp. 1798--1828, 2013.

\bibitem{caljouw2011predictive}
M.~A. Caljouw, W.~P. den Elzen, H.~J. Cools, and J.~Gussekloo, ``Predictive
  factors of urinary tract infections among the oldest old in the general
  population. a population-based prospective follow-up study,'' \emph{BMC
  medicine}, vol.~9, no.~1, p.~57, 2011.

\bibitem{livingston2020dementia}
G.~Livingston, J.~Huntley, A.~Sommerlad, D.~Ames, C.~Ballard, S.~Banerjee,
  C.~Brayne, A.~Burns, J.~Cohen-Mansfield, C.~Cooper \emph{et~al.}, ``Dementia
  prevention, intervention, and care: 2020 report of the lancet commission,''
  \emph{The Lancet}, vol. 396, no. 10248, pp. 413--446, 2020.

\bibitem{pickett2018roadmap}
J.~Pickett, C.~Bird, C.~Ballard, S.~Banerjee, C.~Brayne, K.~Cowan, L.~Clare,
  A.~Comas-Herrera, L.~Corner, S.~Daley \emph{et~al.}, ``A roadmap to advance
  dementia research in prevention, diagnosis, intervention, and care by 2025,''
  \emph{International journal of geriatric psychiatry}, vol.~33, no.~7, pp.
  900--906, 2018.

\end{thebibliography}
\end{document}